\documentclass{article}
\usepackage[preprint,nonatbib]{neurips_data_2024}

\usepackage{hyperref}
\usepackage{url}
\usepackage{booktabs,multirow}
\usepackage{graphicx}
\usepackage{amsthm}
  
\usepackage{amsmath}
\usepackage{amssymb}

\title{Decoupling  Bidirectional Geometric Representations of 4D cost volume with 2D convolution}

\author{\textbf{Xiaobao Wei}\normalfont{\textsuperscript{1,2}}\quad
\textbf{Changyong Shu}\normalfont{\textsuperscript{1}}\footnotemark[1]\quad
\textbf{Zhaokun Yue}\normalfont{\textsuperscript{1}}\quad
\textbf{Chang Huang}\normalfont{\textsuperscript{2}}\quad\\
\textbf{Weiwei Liu}\normalfont{\textsuperscript{2}}\quad
\textbf{Shuai Yang}\normalfont{\textsuperscript{2}}\quad
\textbf{Lirong Yang}\normalfont{\textsuperscript{2}}\quad
\textbf{Peng Gao}\normalfont{\textsuperscript{2}}\quad\\
\textbf{Wenbin Zhang}\normalfont{\textsuperscript{2}}\quad
\textbf{Gaochao Zhu}\normalfont{\textsuperscript{2}}\quad
\textbf{Chengxiang Wang}\normalfont{\textsuperscript{2}} \\
 \\
\textsuperscript{1} Nanjing University of Science and Technology \quad \textsuperscript{2} Carizon \\
\texttt{wxb@njust.edu.cn}
}

\begin{document}

\maketitle
\begin{figure}[htbp]
    \centering
    \includegraphics[width=\linewidth]{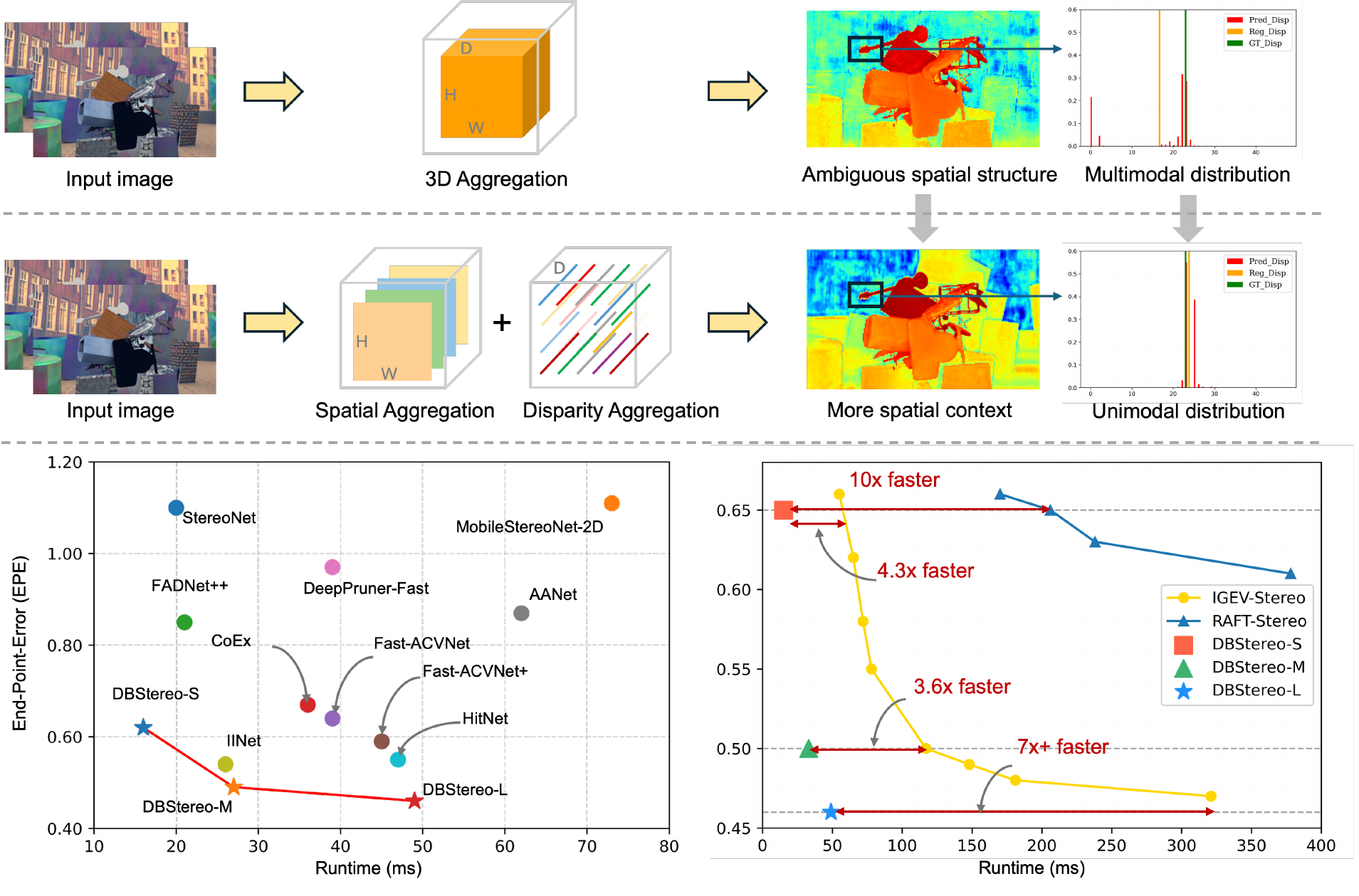}
    \caption{The proposed DBStereo decouple the traditional 3D aggregation into spatial aggregation and disparity aggregation which is based on 2D convolutions. The spatial aggregation can incorporate more spatial structure context and the disparity aggregation make the prediction of disparity more concentrated around the ground truth.
    Our DBStereo outperforms all existing aggregation-based methods~\cite{bangunharcana2021correlate, duggal2019deeppruner,khamis2018stereonet,li2024iinet,Shamsafar_2022_WACV,tankovich2021hitnet,wang2021fadnet++,Xu_2022_CVPR,xu2023accurate,xu2020aanet} in both inference time and accuracy, even surpassing the iterative-based method (RAFT-Stereo~\cite{9665883}and IGEV-Stereo~\cite{Xu_2023_CVPR}).}
    \label{fig:fengmian}
\end{figure}

\begin{abstract}
High-performance real-time stereo matching methods invariably rely on 3D regularization of the cost volume, which is unfriendly to mobile devices.
And 2D regularization based methods struggle in ill-posed regions.
In this paper, we present a deployment-friendly 4D cost aggregation network DBStereo, which is based on pure 2D convolutions. Specifically, we first provide a thorough analysis of the decoupling characteristics of 4D cost volume. And design a lightweight bidirectional geometry aggregation block to capture spatial and disparity representation respectively.
Through decoupled learning, our approach  achieves real-time performance and impressive accuracy simultaneously.
Extensive experiments demonstrate that our proposed DBStereo outperforms all existing aggregation-based methods in both  inference time and accuracy, even surpassing the iterative-based method IGEV-Stereo.
Our study break the empirical design of using 3D convolutions for 4D cost volume and provides a simple yet strong baseline of the proposed decouple aggregation paradigm for further study.
Code will be available at  (\href{https://github.com/happydummy/DBStereo}{https://github.com/happydummy/DBStereo}) soon.
\end{abstract}

\section{Introduction}

Stereo matching has remained a core challenge in computer vision over the past decade, continuously advancing critical applications such as autonomous driving\cite{82}, industrial robotics\cite{81}, and augmented reality\cite{80}. The essence of the technology lies in establishing accurate pixel-level correspondences between left and right images. However, under the resource-constrained conditions of edge computing devices, simultaneously achieving high matching accuracy and real-time inference remains a significant bottleneck.

With the evolution of deep learning, end-to-end stereo matching frameworks have gradually become mainstream. One of the representative works is PSMNet~\cite{Chang_2018_CVPR}, which constructs a 4D cost volume and utilizes 3D convolutional network to aggregate it. Such 4D cost aggregation paradigm methods~\cite{cheng2024adaptive,84,85,86,87} achieve significant breakthroughs on GPU devices. However, the redundant information inherent in 4D cost volumes force the model to rely on computationally expensive 3D convolutions for regularization, posing substantial difficulties for mobile deployment.
In recent years, iterative optimization paradigms~\cite{13,wang2024selective,14,wei2025wavelet}, have demonstrated superior performance. Unlike previous aggregation-based methods, these approaches construct 3D correlation cost volumes and progressively refine disparity maps through iterative indexing it, thereby avoiding complex cost aggregation. While reducing computational complexity, the lack of cost aggregation results in cost volumes deficient in global geometric information, leading to disparity discontinuities in occluded regions, mismatches in textureless areas, and artifacts on reflective surfaces. More critically, achieving acceptable accuracy often requires multiple iterations, resulting in inference delays exceeding 100 ms for most methods, which hinders their applicability in real-time scenarios.

Real-time stereo mathcing research~\cite{bangunharcana2021correlate, duggal2019deeppruner,khamis2018stereonet,li2024iinet,Shamsafar_2022_WACV,tankovich2021hitnet} can be categorized into two types: 2D CNNs based and 3D CNNs based. Both of them made significant compromises: AANet ~\cite{Xu_2020_CVPR} constructs a 3D correlation cost volume and enhances performance in pathological regions by using deformable convolutions, but its specialized operators pose challenges for deployment on edge devices; MobileStereoNet-2D ~\cite{Shamsafar_2022_WACV} attempts a pure 2D convolutional architecture but suffers severe performance degradation; DeepPruner \cite{Duggal_2019_ICCV} narrows the search space by pruning the 4D cost volumes, ACVNet \cite{Xu_2022_CVPR} filters redundant information via attention weights, yet both still rely on 3D CNNs for aggregation. Empirically, it appears that the informative 4D cost volume can not escape its dependence on 3D CNNs.

In fact, these methods overlook inherent limitations of 3D CNNs in stereo matching: spatial and disparity dimensions share the same receptive fields, while disparity aggregation requires a global receptive field, which leading to degradation; the coupled learning of spatial and disparity features increases model training difficulty.
Although FoundationStereo \cite{Wen_2025_CVPR} recognizes the need for different receptive fields of two dimensions and decomposes a 3D convolution into a spatial 3D convolution and a disparity 3D convolution, it remains a localized refinement of standard 3D convolution rather than addressing the fundamental issue of coupled learning.

In this paper,  we propose a novel pure 2D CNN-based framework for 4D cost  aggregation that simultaneously achieves real-time performance and high accuracy.  We first provide an in-depth analysis of the limitations of 3D regularization networks and introduce our spatial-disparity decoupled aggregation paradigm. Specifically, we first use Channel2Disp  operator to transform the 4D cost volume to the 3D one. Then, through our designed Bidirectional Geometry Aggregation (BGA) block consisting of Spatial Aggregation module and  Disparity Aggregation module, we decouple the geometric representation of the cost volume into spatial and disparity dimensions. 
By leveraging 2D CNN-based bidirectional geometric representation decoupling, our method achieves significant improvement. More importantly, our work pioneers a new technical pathway for high-accuracy real-time stereo matching.

Our main contributions are as follows:
\begin{itemize}
    \item We provide a thorough analysis of the geometric representation of 4D cost volume, breaking away from the traditional coupled aggregation paradigm based on 3D convolutions, and establish a simple yet strong baseline for efficient 4D cost aggregation.
    \item We design a pure 2D convolutional Bidirectional Geometry Aggregation  block to independently capture spatial and disparity representation of the 4D cost volume.
    \item We demonstrate the effectiveness of our approach, achieving state-of-the-art performance on multiple benchmarks. The proposed decouple aggregation paradigm opens up a new research direction for the community.
\end{itemize}

\section{Related work}
\label{gen_inst}
Cost aggregation paradigm stereo matching methods~\cite{Duggal_2019_ICCV, Guo_2019_CVPR,Kendall_2017_ICCV,Xu_2022_CVPR,Xu_2020_CVPR} typically follow a four-stage pipeline: feature extraction, cost volume construction, cost aggregation, and disparity regression. Among these, the cost volume serves as the core basis for matching decisions, and its construction quality directly affects final performance.

\textbf{Cost volume construction:} Existing cost volume representation can be divided into two categories: the concatenation volume and the correlation volume. GC-Net~\cite{Kendall_2017_ICCV} directly concatenate the  features maps of left and right images to construct a 4D concatenation cost volume for all disparities. This dense 4D concatenation volume retains comprehensive information from all channels, and thus exhibit enhanced performance. However, excessive redundant information forces the model to rely on a large amount of 3D convolutions to aggregate and regularize the 4D cost volume, which means high computational and memory cost. RAFT-Stereo~\cite{9665883} employs the all-pairs correlation constructed based on a similarity matrix derived from left and right image features.
However just calculates the feature correlation matrix lacks non-local information and struggling in ill-posed regions. GwcNet~\cite{Guo_2019_CVPR} designed a group-wise correlation cost volume that combines the advantage of these two cost volumes. IGEV-Stereo~\cite{Xu_2023_CVPR} constructs a geometry encoding volume incorporating context information and local matching clues.

\textbf{Cost aggregation:} In order to filter the redundant noise on cost volume, cost aggregation consumes a significant amount of computational resources.
DiffuVolume~\cite{zheng2025diffuvolume} design an effective diffusion-based  framework which casts the information filtering as the denoising process of the diffusion model.
ACVNet~\cite{Xu_2022_CVPR} proposed the attention mechanism to filter the cost volume and significantly alleviated the burden of cost aggregation. BANet~\cite{xu2025banet} utlized spatial attention to separate high-frequency edge regions and low-frequency smooth regions of cost volume.
However, these methods still require stacked 3D convolutions to regularize the 4D cost volumes. 
\textbf{Empirically, it seems that high-dimensional cost volume inevitably require dimension-matched convolution to capture the internal correspondences.}

\begin{figure*}[t]
\centering
\vspace{10pt}
\includegraphics[width=\textwidth]{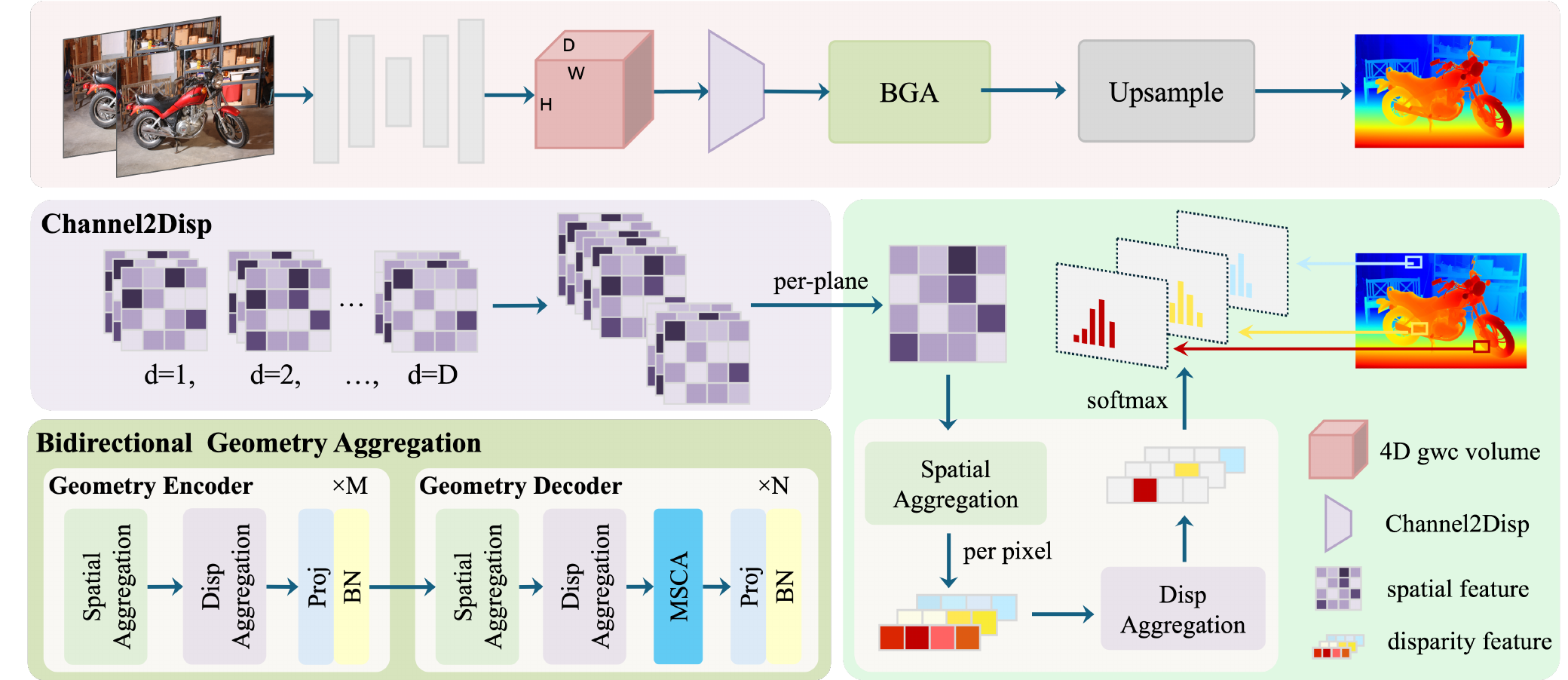}
\caption{The framework of our proposed DBStereo. The Bidirectional
Geometry Aggregation (BGA) block is stacked with multiple
Spatial Aggregation modules and Disparity Aggregation modules. The Spatial Aggregation module extracts spatial context for each  plane of the cost volume and the Disparity Aggregation modules performs global receptive field filtering on each pixel.}
\label{fig:framework}
\end{figure*}

\section{Is 3D Convolution Necessary for 4D Cost Aggregation?}\label{sec:3}
In learning-based stereo matching, constructing a 4D cost volume (of dimensions $(D\times C \times W \times H)$) and applying regularization form the foundation of state-of-the-art paradigms~\cite{Chang_2018_CVPR,Guo_2019_CVPR, Kendall_2017_ICCV,Xu_2022_CVPR}. A widely adopted convention is to directly employ 3D CNNs to process this 4D tensor. The underlying intuition is powerful and seemingly natural: a high-dimensional tensor appears to inherently require convolution operations of matching dimensionality to capture the complex, intertwined relationships across all its dimensions.

However, this empirical design prompts a critical reflection: Is this dimension-matched design truly necessary or optimal? Could it potentially introduce redundancy or even noise? This section delves into the inherent limitations of 3D regularization networks for  stereo matching, thereby motivating our novel  spatial-disparity decoupled aggregation paradigm based on pure 2D convolutions.

\subsection{Limitations of Traditional 3D Regularization Network}
In stereo matching based on 4D cost volumes, traditional aggregation paradigms commonly employ 3D convolutions for cost aggregation. Yet, this paradigm suffers from two inherent flaws:

% \begin{enumerate}
\textbf{Coupled Aggregation Pattern:} The traditional paradigm enforces the use of 3D convolution kernels (e.g., $3\times3\times3$) to extract features from both spatial and disparity dimensions simultaneously.     It implicitly assumes that spatial context and disparity context exhibit consistent aggregation patterns within local regions. However, this assumption is not hold true and could instead constrain the model's expressive capability, making it more susceptible to overfitting noise and non-essential patterns in the training data.

\textbf{Slow Receptive Field Expansion in Disparity Dimension:} Due to the coupled aggregation pattern, the expansion of the receptive field in the disparity dimension is inherently tied to that in the spatial dimensions. Constrained by the kernel size, each 3D convolution operation can only increase the receptive field in the disparity direction marginally (e.g., a $3\times3\times3$ kernel increases it by only 2). To capture sufficient disparity context, a deep stack of 3D CNNs layers is required, directly leading to a dramatic increase in computational cost and memory consumption.
% \end{enumerate}

\subsection{Spatial-Disparity Decoupled Aggregation Paradigm}
\label{sec:innovation}
Based on the above shortcomings, we conducted a thorough analysis of the inherent properties of stereo matching tasks and summarized the following task-specific priors.
% \begin{itemize}

\textbf{Spatial Local Smoothness Prior:} Adjacent pixels at the same depth possess similar disparity values. This prior is particularly beneficial in low-frequency, textureless regions. 
% 3D CNNs couple context from adjacent multiple disparity levels, increasing the difficulty of learning effective spatial representations.

\textbf{Disparity Unimodality Prior:} For a single pixel, the correct disparity value should be unique, meaning the disparity probability distribution should be a sharp unimodal distribution.
% \end{itemize}

We propose a novel spatial-disparity decoupled aggregation paradigm that explicitly encodes these two inherent prior into our network architecture, thereby introducing a powerful inductive bias. Specifically, we reshape the high-dimensional 4D cost volume $(D \times C \times W \times H)$ into a 3D tensor $(D \cdot C \times W \times H)$, decoupling the traditional 4D cost aggregation into two successive pure 2D convolution steps:

% \begin{enumerate}
\textbf{Spatial Aggregation:} A 2D convolution (e.g., with a $3\times3$ kernel) is applied to the spatial dimension of reshaped cost volume $(D\cdot C \times W \times H)$. This step focuses on aggregating spatial context within the same disparity level, effectively smoothing image noise and resolving matching ambiguities in areas like textureless regions.

\textbf{Disparity Aggregation:} Traditional 3D CNNs with their limited local receptive field in disparity dimension struggle to capture long-range dependencies, often resulting in a blurred/tailed distribution or a multi-modal distribution at edges or a flat distribution in textureless areas. We aim for each aggregation along the disparity dimension to possess a global receptive field and apply 2D CNNs with a $1\times1$ kernel to the features after spatial aggregation. It is noteworthy that this $1\times1$ convolution essentially performs a global, fully-connected-style interaction across the entire disparity dimension ($D$), achieving highly efficient optimization of the disparity context.  Through this design, our disparity aggregation module enables comparison and competition across the global disparity range, effectively suppressing incorrect disparity responses and facilitating the formation of a more reasonable, sharper unimodal probability distribution, leading to clearer object boundaries.
% \end{enumerate}

% \subsection{Theoretical Advantages: Strong Inductive Bias and Constrained Hypothesis Space}
% \label{sec:advantages}
% The significant improvements in performance and efficiency observed in our experiments  stem from the fact that our method imposes a much stronger and more reasonable \textit{inductive bias} compared to traditional 3D CNNs, thereby greatly constraining the hypothesis space and achieving superior regularization. The inductive biases introduced by our spatial-disparity decoupled paradigm are:
% \begin{itemize}
%     \item \textbf{Spatial Local Smoothness Prior:} Adjacent pixels at the same depth possess similar disparity values. This bias is particularly beneficial in low-frequency, textureless regions. Traditional 3D convolution couples information from adjacent multiple depths, increasing the difficulty of learning effective spatial aggregation.
%     \item \textbf{Disparity Unimodality Prior:} For a single pixel, the correct disparity value should be unique, meaning the disparity probability distribution should be a sharp unimodal distribution.
% \end{itemize}

\subsection{Conclusion}
\label{sec:conclusion}
In summary, compared to coupled 3D CNNs, our decoupled structure imposes highly precise inductive biases: spatial aggregation enforces \textit{spatial local smoothness prior}, while disparity aggregation enforces \textit{disparity unimodality prior}.  By incorporating these inductive biases into the network architecture, we have significantly reduced the model's search space. The model no longer needs to implicitly learn these fundamental rules from vast amounts of data but instead learns higher level feature representations directly under these inductive biases. This significantly mitigates the risk of the model overfitting to noisy data, thereby achieving a regularization effect far surpassing that of traditional 3D CNNs.

\section{Methods}
\label{headings}
In this section, we introduce the detailed structure of our proposed DBStereo. As shown in Fig \ref{fig:framework}, unlike previous approaches utilizing 3D CNNs, we decouple the 3D regularization network into a bidirectional geometry aggregation module based on purely 2D CNNs:  disparity aggregation module and spatial aggregation module.
\subsection{Feature Extractor}

We employ MobileNetV2 (\cite{Sandler_2018_CVPR}) pretrained on ImageNet (\cite{5206848}) as our backbone to extract multi-scale feature maps $\{\mathbf{f}_{l,i}, \mathbf{f}_{r,i} \in\mathbb{R}^{C_i\times\frac{H}{i}\times\frac{W}{i}}\}, i=4,8,16,32$. And a cascade of upsampling blocks are utilized to restore the feature maps to 1/4 resolution of input image. 
Finally, we obtain multi-scale feature maps $F_{l,i}, F_{r,i} \in\mathbb{R}^{C_i\times\frac{H}{i}\times\frac{W}{i}}\, i=4,8,16$. Among them, $F_{l,4}, F_{r,4}$ are used to construct the 4D cost volume for subsequent disparity prediction,
while $F_{l,4}, F_{l,8}, F_{l,16}$ are utilized to generate spatial attention, further enhancing the robustness of disparity estimation.
\subsection{Cost volume construction}
We construct a 4D group-wise correlation volume~\cite{Guo_2019_CVPR} with features extracted from the left and right images.
The left and right features are split into groups and computing correlation maps group by group.

\begin{align}
\mathbf{C}_{gwc}(d,x,y,g)=\frac{1}{N_c/N_g}\langle\mathbf{f}_l^g(x,y),\mathbf{f}_r^g(x\boldsymbol{-}d,y)\rangle,
\end{align}

where $N_c$ denotes the number of feature channels, $N_g$ denotes the number of groups and $d$ denotes the all disparity candidates.

\subsection{Cost aggregation}
Given the 4D group-wise cost volume, we first use the Channel2Disp operator to fuse the feature dimension and the disparity dimension of the original cost volume.
The core idea of our Channel2Disp transformation is to concatenat the feature maps from all disparity levels, converting the 4D geometric representation into a dense 3D representation without altering spatial structure.
Specifically, we reshape the volume as follows:

\begin{align}
{C_{3D}} = \text{Reshape}({C_{gwc}}) \in \mathbb{R}^{(G \cdot D) \times H \times W}
\end{align}
This operation explicitly encodes the disparity context into a unified  dimension and allows our network to leverage the power of standard 2D convolutions to reason about complex 4D geometric representations without the overhead of 3D operations.

The reconstructed 3D cost volume ${C}_{3D}$ is coupled complex spatial information and disparity information. To efficiently extract required geometric representation, we propose the Bidirectional Geometry Aggregation (BGA) block with encoder-decoder architecture.
The proposal of the BGA is based on the theoretical analysis in Section \ref{sec:3}. We construct the BGA by repeatedly stacking spatial aggregation modules and disparity aggregation modules.

% \textbf{Spatial Structure Aggregation}: We use a series of 2D convolutions with $3\times3$ kernels for spatial structure aggregation. The receptive field of this step is confined to the spatial dimensions $(W, H)$. Its core responsibility is to fuse spatial neighborhood information, enhancing the consistency and smoothness of all disparity features at each location $(x, y)$. 

% \textbf{Disparity Aggregation}: The feature map resulting from spatial aggregation has dimensions $[C_{mid}, W, H]$. We subsequently apply 2D convolutions with $1\times1$ kernels to it. Here, with a $1\times1$ receptive field, it ceases to fuse spatial information and instead operates along the \textbf{channel dimension}. Since the channel dimension is a recombination of the original disparity $D$ and feature $C$, this operation inherently \textbf{learns the complex relationships between different disparity hypotheses implicitly, \textit{within} each spatial location $(x, y)$}, facilitating competition and selection among disparities. 

\subsection{Disparity Prediction}
After obtaining the aggregated cost volume, we apply the softmax operation to it to regress the disparity map $d_0$:
\begin{align}
\mathbf{d}_0 = \sum_{d=0}^{D_{max}/4-1} d \times \text{Softmax}(\mathbf{C}_{agg}(d)),
\end{align}
where $D_{max}$ denotes the predefined maximum disparity. The disparity map $d_0$ is at $1/4$ resolution of input images. We utilize interpolation and learnable parameters respectively to upsample the disparity map $d_0$ to full resolution for supervision.

\subsection{Loss Function}

We employ the smooth $L_1$ loss to supervise our network. The loss is defined as follow:

\begin{align}
\mathcal{L}=\lambda_0Smooth_{L_1}(\mathbf{d}_{init}-\mathbf{d}_{gt})+\lambda_1Smooth_{L_1}(\mathbf{d}_{final}-\mathbf{d}_{gt})
\end{align} 
where $d_{gt}$ is the ground truth of disparity and $\lambda_0 = 0.3, \lambda_1 = 1$.

% \begin{table}[t]
%   \centering
%   \caption{Comparison with the state-of-the-art on SceneFlow. Runtime is measured on an RTX~3090 GPU.}
%   \label{tab:sceneflow}
%   \renewcommand{\arraystretch}{1.1}
%   \setlength{\tabcolsep}{4.5pt}
%  \begin{tabular}{c|cccc}
%   \toprule[1.5pt]
%   \multicolumn{1}{c|}{\textbf{Paradigm}} &
%   \textbf{Method} & \textbf{Params (M)} & \textbf{EPE (px)} & \textbf{Runtime (ms)} \\
%   \midrule
%   \multirow{12}{*}{\parbox{2.5cm}{\centering Cost   aggregation}}
%   &DeepPruner-Fast      & 7.47  & 1.25 & 62 \\
%   &StereoNet            & 0.40  & 1.10 & 20 \\
%   &AANet                & 2.97  & 0.87 & 93 \\
%   &MobileStereoNet-2D   & 2.23  & 1.11 & 73 \\
%   &MobileStereoNet-3D   & 2.23  & 0.80 & 73 \\
%   &FADNet++             & 124.26& 0.85 & 21 \\
%   &CoEx                 & 2.72  & 0.67 & 36 \\
%   &Fast-ACVNet          & 3.08  & 0.64 & 22 \\
%   &ACVNet               & 3.20  & 0.59 & 27 \\
%   &IINet                & 19.54 & 0.54 & 26 \\
%   &BANET-2D             & 5.46    & 0.57 & 41\\
%   &BANET-3D             & 3.63    & 0.51 & 105\\
%   \midrule
%   \multirow{2}{*}{\parbox{2.5cm}{\centering Iterative optimization}}
%   &RAFT-Stereo          & --    & 0.61 & 380\\[2pt]
%   &IGEV-Stereo          & 12.6    & \textbf{0.47} & 340\\[2pt]
%   \midrule
%   \multirow{3}{*}{\parbox{2.5cm}{\centering Decouple aggregation}}
%   &DBStereo-S (Ours) & 2.57 & 0.65 & \textbf{15} \\
%   &DBStereo-M (Ours) & 30.2 & 0.50 & 33 \\
%   &DBStereo-L (Ours)    & 68.49& \textbf{0.46} & 49 \\
%   \bottomrule[1.5pt]
% \end{tabular}
% \end{table}
\begin{table}[t]
  \centering
  \caption{Comparison with the state-of-the-art methods on SceneFlow. Runtime is measured on an RTX~3090 GPU.}
  \label{tab:sceneflow}
  \renewcommand{\arraystretch}{1.1}
  \setlength{\tabcolsep}{4.5pt}
 \begin{tabular}{c|cccc}
  \toprule[1.5pt]
  \multicolumn{1}{c|}{\textbf{Paradigm}} &
  \textbf{Method}  & \textbf{EPE (px)} & \textbf{D1 (\%)}& \textbf{Runtime (ms)} \\
  \midrule
  \multirow{12}{*}{\parbox{2.5cm}{\centering Cost   aggregation}}
  &PSMNet~\cite{Chang_2018_CVPR}        & 1.09 & 12.1& 317 \\
  &StereoNet~\cite{khamis2018stereonet}              & 1.10 & - & 20 \\
  &AANet~\cite{Xu_2020_CVPR}                  & 0.87 & 9.3& 93 \\
  &AANet+~\cite{Xu_2020_CVPR}                   & 0.72 & 7.4 & 87 \\
  &MobileStereoNet-2D~\cite{Shamsafar_2022_WACV}     & 1.11 & - & 73 \\
  % &MobileStereoNet-3D     & 0.80 && 73 \\
  &FADNet++~\cite{wang2021fadnet++}               & 0.85 & - & 21 \\
  &CoEx~\cite{bangunharcana2021correlate}                  & 0.67 & 4.02 & 36 \\
  &Fast-ACVNet~\cite{xu2023accurate}            & 0.64 & 2.31 & 39 \\
  &Fast-ACVNet+~\cite{xu2023accurate}           & 0.59 & 2.08 & 45 \\
  &IINet~\cite{li2024iinet}                  & 0.54 & 2.18 & 26 \\
  &BANET-2D~\cite{xu2025banet}               & 0.57 & 2.50 & xx\\
  &BANET-3D~\cite{xu2025banet}               & 0.51 & 2.21 & xx\\
  \midrule
  \multirow{2}{*}{\parbox{2.5cm}{\centering Iterative optimization}}
  &RAFT-Stereo~\cite{9665883}          & 0.61    & 2.85 & 380\\[2pt]
  &IGEV-Stereo~\cite{Xu_2023_CVPR}          & \underline{0.47}    & 2.47 & 340\\[2pt]
  \midrule
  \multirow{3}{*}{\parbox{2.5cm}{\centering Decouple aggregation}}
  &DBStereo-S (Ours) & 0.65 & 2.36 & \textbf{15} \\
  &DBStereo-M (Ours) & 0.50 & 1.80 & 33 \\
  &DBStereo-L (Ours)    & \textbf{0.45}& \textbf{1.57} & 49 \\
  \bottomrule[1.5pt]
\end{tabular}
\end{table}

\section{Experiments}
\label{others}
\subsection{Datasets and Evaluation Metrics}
\textbf{Scene Flow~\cite{27}} is a large-scale synthetic stereo dataset containing 35,454 training and 4,370 testing stereo image pairs at 960×540 resolution. This datasets provide dense disparity map as ground truth. In evaluations, we utilize the end point error (EPE) and the D1 outlier as the evaluation metrics, where EPE is the average $l_{1}$ distance between the prediction and ground truth disparity. And D1 denotes the percentage of outliers with an absolute error greater than 1 pixels.

\textbf{KITTI} is a real-world dataset consisting of KITTI 2012~\cite{28} and KITTI 2015~\cite{29}. KITTI 2012 provides 194 training pairs and 195 testing pairs, and KITTI 2015 provides 200 training pairs and 200 testing pairs. Both datasets provide sparse ground-truth disparities obtained with LiDAR. For evaluations, we calculate EPE and the percentage of pixels with EPE larger than 3 pixels in all (D1-all) regions. All two KITTI datasets are also used for cross-domain generalization performance evaluation, with EPE and $>$3px metric (i.e., the percentage of points with absolute error larger than 3 pixels) reported.

\subsection{Implemention Details}
We have implemented our methods using PyTorch and conducted experiments on 8 NVIDIA RTX 3090 GPUs. We train our pretrained model on Scene Flow dataset for 90 epochs. For the KITTI dataset evaluation, we fine-tuned the pre-trained model for 500 epochs using a mixed training set comprising KITTI 2012 and KITTI2015 training datasets.

\subsection{Benchmark datasets and Performance}
We evaluate our DBStereo on two widely used benchmarks and submit the results to online
leaderboards for public comparison: Scene Flow, KITTI 2012 and KITTI 2015.

\textbf{Scene Flow:}  As shown in Table \ref{tab:sceneflow}, we compare our proposed DBStereo and its variances  with several state-of-the-art approaches on the SceneFlow dataset. Our DBStereo-L achieves the highest accuracy among all the published real-time methods and even surpass many high-performance iterative-based methods both accuracy and inference time such as RAFT-Stereo and IGEV-Sterero, reducing the runtime by more than 85\%.

\section{Conclusion}
In this paper, we provide a thorough analysis of the limitations of traditional aggregation paradigm methods, breaking the empirical approach of using dimension-matched convolutions for a high-dimensional cost volume. We propose the DBStereo which is based on pure 2D convolutions but achieve impressive performance both in accuracy and inferenc time.
DBStereo is a simple yet strong baseline of our proposed decouple aggregation paradigm. We hope our research will provide some insightful directions for future community studies.

\bibliography{main}

\begin{thebibliography}{10}

\bibitem{bangunharcana2021correlate}
A.~Bangunharcana, J.~W. Cho, S.~Lee, I.~S. Kweon, K.-S. Kim, and S.~Kim.
\newblock Correlate-and-excite: Real-time stereo matching via guided cost volume excitation.
\newblock In {\em 2021 IEEE/RSJ International Conference on Intelligent Robots and Systems (IROS)}, pages 3542--3548. IEEE, 2021.

\bibitem{Chang_2018_CVPR}
J.-R. Chang and Y.-S. Chen.
\newblock Pyramid stereo matching network.
\newblock In {\em Proceedings of the IEEE Conference on Computer Vision and Pattern Recognition (CVPR)}, June 2018.

\bibitem{cheng2024adaptive}
J.~Cheng, W.~Yin, K.~Wang, X.~Chen, S.~Wang, and X.~Yang.
\newblock Adaptive fusion of single-view and multi-view depth for autonomous driving.
\newblock In {\em Proceedings of the IEEE/CVF Conference on Computer Vision and Pattern Recognition}, pages 10138--10147, 2024.

\bibitem{5206848}
J.~Deng, W.~Dong, R.~Socher, L.-J. Li, K.~Li, and L.~Fei-Fei.
\newblock Imagenet: A large-scale hierarchical image database.
\newblock In {\em 2009 IEEE Conference on Computer Vision and Pattern Recognition}, pages 248--255, 2009.

\bibitem{duggal2019deeppruner}
S.~Duggal, S.~Wang, W.-C. Ma, R.~Hu, and R.~Urtasun.
\newblock Deeppruner: Learning efficient stereo matching via differentiable patchmatch.
\newblock In {\em Proceedings of the IEEE/CVF international conference on computer vision}, pages 4384--4393, 2019.

\bibitem{84}
S.~Duggal, S.~Wang, W.-C. Ma, R.~Hu, and R.~Urtasun.
\newblock Deeppruner: Learning efficient stereo matching via differentiable patchmatch.
\newblock In {\em Proceedings of the IEEE/CVF International Conference on Computer Vision}, pages 4384--4393, 2019.

\bibitem{Duggal_2019_ICCV}
S.~Duggal, S.~Wang, W.-C. Ma, R.~Hu, and R.~Urtasun.
\newblock Deeppruner: Learning efficient stereo matching via differentiable patchmatch.
\newblock In {\em Proceedings of the IEEE/CVF International Conference on Computer Vision (ICCV)}, October 2019.

\bibitem{28}
A.~Geiger, P.~Lenz, and R.~Urtasun.
\newblock Are we ready for autonomous driving? the kitti vision benchmark suite.
\newblock In {\em 2012 IEEE Conference on Computer Vision and Pattern Recognition}, pages 3354--3361. IEEE, 2012.

\bibitem{Guo_2019_CVPR}
X.~Guo, K.~Yang, W.~Yang, X.~Wang, and H.~Li.
\newblock Group-wise correlation stereo network.
\newblock In {\em Proceedings of the IEEE/CVF Conference on Computer Vision and Pattern Recognition (CVPR)}, June 2019.

\bibitem{81}
Y.-Z. Hsieh and S.-S. Lin.
\newblock Robotic arm assistance system based on simple stereo matching and q-learning optimization.
\newblock {\em IEEE Sensors Journal}, 20(18):10945--10954, 2020.

\bibitem{Kendall_2017_ICCV}
A.~Kendall, H.~Martirosyan, S.~Dasgupta, P.~Henry, R.~Kennedy, A.~Bachrach, and A.~Bry.
\newblock End-to-end learning of geometry and context for deep stereo regression.
\newblock In {\em Proceedings of the IEEE International Conference on Computer Vision (ICCV)}, Oct 2017.

\bibitem{khamis2018stereonet}
S.~Khamis, S.~Fanello, C.~Rhemann, A.~Kowdle, J.~Valentin, and S.~Izadi.
\newblock Stereonet: Guided hierarchical refinement for real-time edge-aware depth prediction.
\newblock In {\em Proceedings of the European conference on computer vision (ECCV)}, pages 573--590, 2018.

\bibitem{li2024iinet}
X.~Li, C.~Zhang, W.~Su, and W.~Tao.
\newblock Iinet: Implicit intra-inter information fusion for real-time stereo matching.
\newblock In {\em Proceedings of the AAAI Conference on Artificial Intelligence}, volume~38, pages 3225--3233, 2024.

\bibitem{85}
Z.~Liang, Y.~Guo, Y.~Feng, W.~Chen, L.~Qiao, L.~Zhou, J.~Zhang, and H.~Liu.
\newblock Stereo matching using multi-level cost volume and multi-scale feature constancy.
\newblock {\em IEEE Transactions on Pattern Analysis and Machine Intelligence}, 43(1):300--315, 2019.

\bibitem{9665883}
L.~Lipson, Z.~Teed, and J.~Deng.
\newblock Raft-stereo: Multilevel recurrent field transforms for stereo matching.
\newblock In {\em 2021 International Conference on 3D Vision (3DV)}, pages 218--227, 2021.

\bibitem{13}
L.~Lipson, Z.~Teed, and J.~Deng.
\newblock Raft-stereo: Multilevel recurrent field transforms for stereo matching.
\newblock In {\em 2021 International Conference on 3D Vision (3DV)}, pages 218--227. IEEE, 2021.

\bibitem{27}
N.~Mayer, E.~Ilg, P.~Hausser, P.~Fischer, D.~Cremers, A.~Dosovitskiy, and T.~Brox.
\newblock A large dataset to train convolutional networks for disparity, optical flow, and scene flow estimation.
\newblock In {\em Proceedings of the IEEE Conference on Computer Vision and Pattern Recognition}, pages 4040--4048, 2016.

\bibitem{29}
M.~Menze and A.~Geiger.
\newblock Object scene flow for autonomous vehicles.
\newblock In {\em Proceedings of the IEEE Conference on Computer Vision and Pattern Recognition}, pages 3061--3070, 2015.

\bibitem{86}
G.-Y. Nie, M.-M. Cheng, Y.~Liu, Z.~Liang, D.-P. Fan, Y.~Liu, and Y.~Wang.
\newblock Multi-level context ultra-aggregation for stereo matching.
\newblock In {\em Proceedings of the IEEE/CVF Conference on Computer Vision and Pattern Recognition}, pages 3283--3291, 2019.

\bibitem{Sandler_2018_CVPR}
M.~Sandler, A.~Howard, M.~Zhu, A.~Zhmoginov, and L.-C. Chen.
\newblock Mobilenetv2: Inverted residuals and linear bottlenecks.
\newblock In {\em Proceedings of the IEEE Conference on Computer Vision and Pattern Recognition (CVPR)}, June 2018.

\bibitem{Shamsafar_2022_WACV}
F.~Shamsafar, S.~Woerz, R.~Rahim, and A.~Zell.
\newblock Mobilestereonet: Towards lightweight deep networks for stereo matching.
\newblock In {\em Proceedings of the IEEE/CVF Winter Conference on Applications of Computer Vision (WACV)}, pages 2417--2426, January 2022.

\bibitem{tankovich2021hitnet}
V.~Tankovich, C.~Hane, Y.~Zhang, A.~Kowdle, S.~Fanello, and S.~Bouaziz.
\newblock Hitnet: Hierarchical iterative tile refinement network for real-time stereo matching.
\newblock In {\em Proceedings of the IEEE/CVF conference on computer vision and pattern recognition}, pages 14362--14372, 2021.

\bibitem{wang2021fadnet++}
Q.~Wang, S.~Shi, S.~Zheng, K.~Zhao, and X.~Chu.
\newblock Fadnet++: Real-time and accurate disparity estimation with configurable networks.
\newblock {\em arXiv preprint arXiv:2110.02582}, 2021.

\bibitem{wang2024selective}
X.~Wang, G.~Xu, H.~Jia, and X.~Yang.
\newblock Selective-stereo: Adaptive frequency information selection for stereo matching.
\newblock In {\em Proceedings of the IEEE/CVF Conference on Computer Vision and Pattern Recognition}, pages 19701--19710, 2024.

\bibitem{wei2025wavelet}
X.~Wei, J.~Liu, D.~Yang, J.~Cheng, C.~Shu, and W.~Wang.
\newblock A wavelet-based stereo matching framework for solving frequency convergence inconsistency.
\newblock {\em arXiv preprint arXiv:2505.18024}, 2025.

\bibitem{Wen_2025_CVPR}
B.~Wen, M.~Trepte, J.~Aribido, J.~Kautz, O.~Gallo, and S.~Birchfield.
\newblock Foundationstereo: Zero-shot stereo matching.
\newblock In {\em Proceedings of the IEEE/CVF Conference on Computer Vision and Pattern Recognition (CVPR)}, pages 5249--5260, June 2025.

\bibitem{87}
Z.~Wu, X.~Wu, X.~Zhang, S.~Wang, and L.~Ju.
\newblock Semantic stereo matching with pyramid cost volumes.
\newblock In {\em Proceedings of the IEEE/CVF International Conference on Computer Vision}, pages 7484--7493, 2019.

\bibitem{Xu_2022_CVPR}
G.~Xu, J.~Cheng, P.~Guo, and X.~Yang.
\newblock Attention concatenation volume for accurate and efficient stereo matching.
\newblock In {\em Proceedings of the IEEE/CVF Conference on Computer Vision and Pattern Recognition (CVPR)}, pages 12981--12990, June 2022.

\bibitem{xu2025banet}
G.~Xu, J.~Liu, X.~Wang, J.~Cheng, Y.~Deng, J.~Zang, Y.~Chen, and X.~Yang.
\newblock Banet: Bilateral aggregation network for mobile stereo matching.
\newblock {\em arXiv preprint arXiv:2503.03259}, 2025.

\bibitem{Xu_2023_CVPR}
G.~Xu, X.~Wang, X.~Ding, and X.~Yang.
\newblock Iterative geometry encoding volume for stereo matching.
\newblock In {\em Proceedings of the IEEE/CVF Conference on Computer Vision and Pattern Recognition (CVPR)}, pages 21919--21928, June 2023.

\bibitem{14}
G.~Xu, X.~Wang, X.~Ding, and X.~Yang.
\newblock Iterative geometry encoding volume for stereo matching.
\newblock In {\em Proceedings of the IEEE/CVF Conference on Computer Vision and Pattern Recognition}, pages 21919--21928, 2023.

\bibitem{xu2023accurate}
G.~Xu, Y.~Wang, J.~Cheng, J.~Tang, and X.~Yang.
\newblock Accurate and efficient stereo matching via attention concatenation volume.
\newblock {\em IEEE Transactions on Pattern Analysis and Machine Intelligence}, 46(4):2461--2474, 2023.

\bibitem{xu2020aanet}
H.~Xu and J.~Zhang.
\newblock Aanet: Adaptive aggregation network for efficient stereo matching.
\newblock In {\em Proceedings of the IEEE/CVF conference on computer vision and pattern recognition}, pages 1959--1968, 2020.

\bibitem{Xu_2020_CVPR}
H.~Xu and J.~Zhang.
\newblock Aanet: Adaptive aggregation network for efficient stereo matching.
\newblock In {\em Proceedings of the IEEE/CVF Conference on Computer Vision and Pattern Recognition (CVPR)}, June 2020.

\bibitem{82}
G.~Yang, X.~Song, C.~Huang, Z.~Deng, J.~Shi, and B.~Zhou.
\newblock Drivingstereo: A large-scale dataset for stereo matching in autonomous driving scenarios.
\newblock In {\em Proceedings of the IEEE/CVF Conference on Computer Vision and Pattern Recognition}, pages 899--908, 2019.

\bibitem{80}
N.~Zenati and N.~Zerhouni.
\newblock Dense stereo matching with application to augmented reality.
\newblock In {\em 2007 IEEE International Conference on Signal Processing and Communications}, pages 1503--1506. IEEE, 2007.

\bibitem{zheng2025diffuvolume}
D.~Zheng, X.-M. Wu, Z.~Liu, J.~Meng, and W.-s. Zheng.
\newblock Diffuvolume: Diffusion model for volume based stereo matching.
\newblock {\em International Journal of Computer Vision}, 133(7):3807--3821, 2025.

\end{thebibliography}
\bibliographystyle{abbrv}
% \appendix
% \section{Appendix}

\end{document}